\documentclass{article} 
\usepackage{iclr2021_conference,times}

\usepackage{hyperref}
\usepackage{url}
\usepackage[utf8]{inputenc} 
\usepackage[T1]{fontenc}    
\usepackage{hyperref}       
\usepackage{url}            
\usepackage{booktabs}       
\usepackage{amsfonts,amsmath}       
\usepackage{nicefrac}       
\usepackage{microtype}      
\usepackage{color}
\usepackage{caption}
\usepackage{subcaption}
\usepackage{graphicx}
\usepackage{authblk}
\usepackage{adjustbox}
\usepackage{color}

\title{Simple Transparent Adversarial Examples}

\author[ ~1, 2]{Jaydeep Borkar\thanks{Work done while Jaydeep was an external student at MIT-IBM Watson AI Lab, IBM Research.}}
\author[1]{Pin-Yu Chen}
\affil[1]{MIT-IBM Watson AI Lab, IBM Research}
\affil[2]{University of Pune}
\affil[ ]{\{jaijborkar@gmail.com, pin-yu.chen@ibm.com\}}

\iclrfinalcopy  
\begin{document}

\maketitle

\begin{abstract}
There has been a rise in the use of Machine Learning as a Service (MLaaS) Vision APIs as they offer multiple services including pre-built models and algorithms, which otherwise take a huge amount of resources if built from scratch. As these APIs get deployed for high-stakes applications, it’s very important that they are robust to different manipulations. Recent works have only focused on typical adversarial attacks when evaluating the robustness of vision APIs. We propose two new aspects of adversarial image generation methods and evaluate them on the robustness of Google Cloud Vision API’s optical character recognition service and object detection APIs deployed in real-world settings such as sightengine.com, picpurify.com, Google Cloud Vision API, and Microsoft Azure's computer vision API. Specifically, we go beyond the conventional ``small-noise'' adversarial attacks and introduce \textit{secret embedding} and \textit{transparent adversarial examples} as a simpler way to evaluate robustness. These methods are so straightforward that even non-specialists can craft such attacks. As a result, they pose a serious threat where APIs are used for high-stakes applications. Our transparent adversarial examples successfully evade state-of-the-art object detection APIs such as Azure Cloud Vision (attack success rate 52\%) and Google Cloud Vision (attack success rate 36\%). 90\% of the images have a  secret embedded text that successfully fools the vision of time-limited humans but is detected by Google Cloud Vision API's optical character recognition. Complementing to current research, our results provide simple but  unconventional methods on robustness evaluation.  
\end{abstract}

\section{Introduction}

Deep neural networks have been found to be vulnerable to adversarial attacks \citep{42503, Biggio_2013,43405}, where it is possible to add 
perturbations to the image that results in misclassification. These perturbations are such that the perturbed image still looks semantically similar to the original image to humans. These perturbed samples are called as adversarial examples. The current methods for generating adversarial images require the attacker to have specific knowledge about the victim model (e.g. the input gradient used in white-box attack) or excessive number of prediction evaluations (e.g. gradient estimation or substitute model training with model queries in black-box attacks). They may not closely resemble the real-world attack scenarios where the attacker might not have machine learning and programming skills to carefully craft adversarial examples and the budget to carry out computationally expensive black-box attacks. 

In this work, we argue that though the current approaches of generating adversarial examples are important and well studied with respect to a defined threat model (e.g. $\ell_p$ norm bounded perturbation), they do not provide a complete spectrum of robustness evaluation using semantically similar adversarial examples. Specifically, we propose new and straightforward manipulations for Optical Character Recognition (OCR) and object detection APIs under a new type of simple black-box attack called \textit{Simple Transparent Adversarial Examples}, where anyone without any machine learning expertise can very easily create simple prediction-evasive and semantically similar examples and successfully fool deep neural networks powered vision APIs. They also do not need any computational power and so they are not computationally expensive contrary to the current adversarial examples generation methods. Therefore, our simple attack is more realistic in real-world settings considering the safety of publicly deployed AI systems and potentially more dangerous than the current black-box attacks.  

We also show that our attack is adversarial in two folds: (1) It's prediction evasion (i.e. the transparent adversarial examples evade object detection of the computer vision APIs) (2) The adversarial images generated by secret embedding approach carry information (text) that is detected by Google Cloud Vision API but evades the vision of time-limited humans.

\section{Simple Transparent Adversarial Examples}

The current methods of generating adversarial examples require the attacker to have a certain level of machine learning expertise (e.g. first-order or zeroth-order optimization for handling adversarial attacks with perturbation constraints) to craft these attacks \citep{42503, 43405, madry2019deep, 7467366, 46561, carlini2017towards, 10.1145/3052973.3053009, CPY17zoo, hosseini2018semantic}. Thus, only the attacker who has the required domain knowledge can attack deep neural networks. We propose a new type of simple attack that doesn't require the attacker to have any machine learning and programming knowledge to attack deep neural networks based cloud vision APIs. We talk about how our attack is different than some of the recently proposed simple manipulations in \ref{appendix:simple} 

To demonstrate this attack, we focus on attacking machine learning as a service (MLaaS) cloud vision APIs such as sightengine.com\footnote{\url{https://sightengine.com/detect-weapons-alcohol-drugs}}, PicPurify\footnote{\url{https://www.picpurify.com/demo-gun.html}}, Google Cloud Vision API\footnote{\url{https://cloud.google.com/vision}}, and Microsoft Azure's computer vision API\footnote{\url{https://azure.microsoft.com/en-in/services/cognitive-services/computer-vision/}}. We query these APIs via the web interface available through the demo. We test our attack on two of the most popular services offered by these visions APIs: optical character recognition (by Google Cloud Vision API) and object detection. \textit{Simple Transparent Adversarial Examples} have dual definitions in our work: (1) Examples that are straightforward to craft and don't need any machine learning and programming knowledge (i.e. simple and transparent to craft) and (2) Examples modified by introducing transparent white regions and secret invisible text embedding.

\subsection{Simple Transparent Adversarial Examples to Evade Object Detection}

Recent black-box adversarial attack methods \citep{10.1145/3052973.3053009, CPY17zoo} and adversarial attack methods for object detectors and Google Cloud Vision API \citep{liu2019dpatch, li2019exploring, Chen_2019, 220580, hosseini2017googles, goodman2020transferability} are small noise adversarial attacks that require the attacker to have machine learning and programming knowledge to carefully design such attacks. We propose a simple attack method where the attacker can add transparent white patches to an image using any publicly available online transparency tool to successfully evade state-of-the-art object detection APIs or misclassify the labels. Since anyone even without any machine learning and programming knowledge can easily fool the publicly deployed APIs, these type of attacks pose a higher security risk than the current typical adversarial attacks. 

Moreover, the current black-box adversarial attack methods \citep{10.1145/3052973.3053009} require thousands of queries to design adversarial examples that can be very expensive. In our proposed attack, the attacker can successfully attack object detection APIs with just a few queries (See Table \ref{table:1}). Hence, we argue that our black-box attack is cheaper and query efficient than the majority of current black-box attack methods with a different modification (perturbation) constraint.

\subsubsection{Attack creation}
For any image \textit{$x_0$}, we perturb \textit{$x_0$} by introducing a white transparent patch \textit{p} with a modification constraint $\epsilon$, such that the resultant modified image \textit{x} is either misclassified or evades the classification entirely. The value of $\epsilon$ is such that the modified image still remains unambiguous and class-preserving to humans. We use publicly available onlinejpgtools.com's transparency maker tool \footnote{\url{https://onlinejpgtools.com/make-jpg-transparent}} to introduce white transparent patch in the image. We use a publicly available online tool to demonstrate that anyone can easily craft such examples to attack publicly deployed APIs, hence these type of attacks pose a higher security risk than the typical adversarial attacks where the attacker is required to have machine learning expertise to design attacks. We use Kaggle's Weapons Dataset \footnote{\url{https://www.kaggle.com/ar5p1edy/weapons-datasets}} to generate adversarial images and test them on weapon detection APIs (PicPurify and Sightengine.com) and general object detection APIs (Azure vision and Google Cloud Vision). For adding the transparent patch on the image, we follow two approaches: 1) We don't select the region(s) to be patched ourselves (i.e. we let the tool decide it) 2) We randomly select the region to be patched. We propose attack algorithms for both the approaches in \ref{appendix: attack1}.

\subsubsection{Experiments and Observations}
\begin{figure} [!hbt]  
  \begin{subfigure}{0.48\textwidth}
    \includegraphics[width=\linewidth]{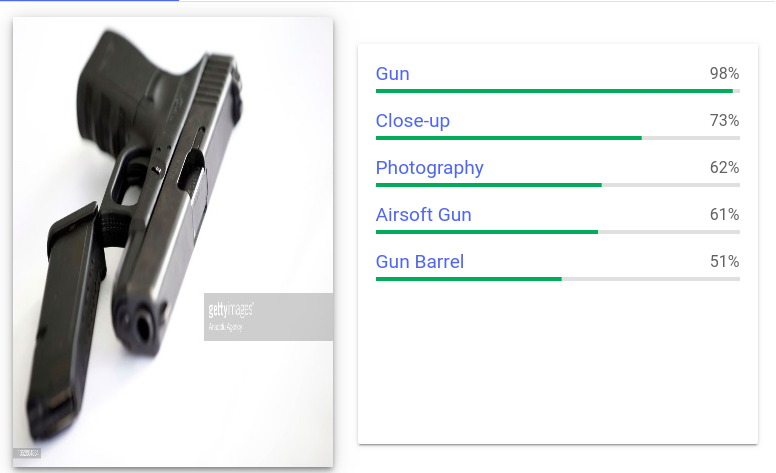}
    \caption{Original image} 
  \end{subfigure}\hspace{0.5em}%
  \begin{subfigure}{0.48\textwidth}
    \includegraphics[width=\linewidth]{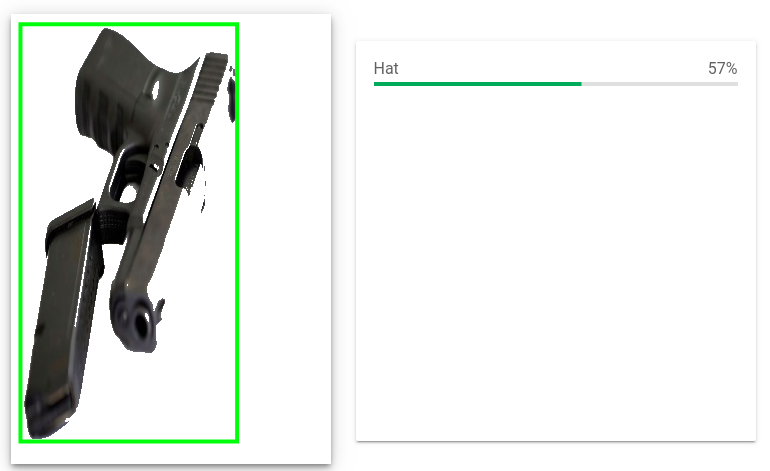}
    \caption{Modified image} 
  \end{subfigure} 
  \caption{(a) The original image that gets detected successfully by Google Cloud Vision API. (b) Modified image with 55 $\%$ transparency intensity getting misclassified as a \textbf{hat} by Google Cloud Vision API.} \label{fig:1}
  \vspace{-4mm}
\end{figure}

We find that Simple Transparent Adversarial Examples successfully evade the weapon/object detection by sightengine.com, PicPurify, Google Cloud Vision API and Microsoft Azure's computer vision API. Table \ref{table:1} in \ref{appendix: performance2} shows the performance of each API against these examples. Figure \ref{fig:1} illustrates this attack. 

\subsection{Simple Transparent Adversarial Examples for Optical Character Recognition (OCR)}
We propose a new method called \textit{secret embedding approach} to modify images by embedding text in them. We call such images as \textit{Simple Transparent Adversarial Examples} for OCR as they are straightforward to craft and do not require any machine learning or programming skills. We evaluate these images on Google Cloud Vision API's OCR feature\footnote{\url{https://cloud.google.com/vision/docs/ocr}}. We find that the images created by this method carry information (text) that is only recognizable by Google Cloud Vision API's OCR whereas it evades the vision of time-limited humans. In this sense, they go beyond the traditional definition of adversarial examples where they fool deep neural networks and not humans. The closest line of work to this is by \citep{NIPS2018_7647}, where the adversarial examples fool both computer vision and time-limited humans. We find that our examples also successfully evade the vision of both computer vision (i.e. Google Cloud Vision API) and time-limited humans at a specific font size and color (see \ref{appendix:foolboth}).

\subsubsection{Attack creation}
For an image $\textit{x}_0$, we find an image \textit{x} with an embedded text \textit{t}, such that \textit{t} evades the vision of time-limited humans but is detected by Google Cloud Vision's Optical Character Recognition (OCR). To embed the text, we use an online tool imagecolorpicker.com \footnote{\url{imagecolorpicker.com/en/}} to find RGB values of the region where we want to embed the text. Next, we embed the text using an online editing tool Photopea \footnote{\url{https://www.photopea.com/}}. We use publicly available online tools to show that such attacks can be easily designed by anyone, even without any machine learning and programming knowledge. We use images from Caltech 101 \footnote{\url{http://www.vision.caltech.edu/Image_Datasets/Caltech101/}} and Caltech-256 \footnote{\url{https://authors.library.caltech.edu/7694/}} dataset for this attack. We propose a simple algorithm for \textit{secret embedding approach} in \ref{appendix: ocrattackalgo}.

\subsubsection{Observations}
\begin{figure}[hbt!] 
  \begin{subfigure}{0.23\textwidth}
    \includegraphics[width=\linewidth]{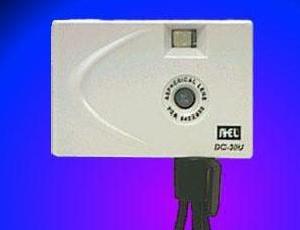}
    \caption{Original image} 
  \end{subfigure}\hspace{1em}%
  \begin{subfigure}{0.23\textwidth}
    \includegraphics[width=\linewidth]{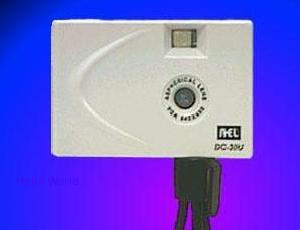}
    \caption{Modified image} 
  \end{subfigure}\hspace{1em}%
  \begin{subfigure}{0.23\textwidth}
    \includegraphics[width=\linewidth]{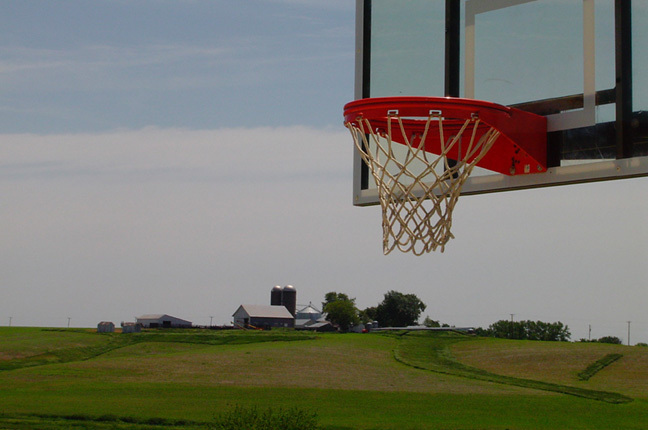}
    \caption{Original image} 
  \end{subfigure}\hspace{1em}%
   \begin{subfigure}{0.23\textwidth}
    \includegraphics[width=\linewidth]{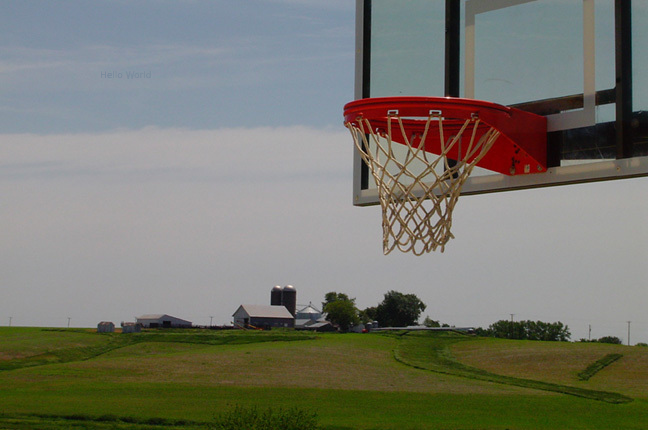}
    \caption{Modified image} 
  \end{subfigure}
\caption{(b) \& (d) are modified images that fool time-limited humans but not Google Cloud Vision's OCR. (b) has an embedded text "Hello World" of font size 11 px and RGB difference 30 (d) has an embedded text "Hello World" of font size 9 px and RGB difference 30. } \label{fig:2}
\end{figure}

We find that in some cases there is a trade-off between the font size of the embedded text and the RGB difference\footnote{RGB difference defined in \ref{appendix: ocrattackalgo}}. To elaborate, text with higher RGB difference and smaller font size ($\leq$15 px) or text with smaller RGB difference and relatively larger font size (>15 px) is favorable to craft these examples. However, smaller RGB difference and smaller font size are the ideal conditions. Figure \ref{fig:2} illustrates this attack. In our experiments we test 40 images. 90\% of those images fooled the vision of time-limited humans and were recognized by Google Cloud Vision OCR. However, we also achieved near 100\% evasion rate in the case of examples that fooled both time-limited humans and OCR with the secret text of very small font size and RGB difference set to 0 (For example see \ref{appendix:foolboth}). The importance of this attack and it's possible risks and applications are highlighted in \ref{appendix: importance} and \ref{appendix: risks} in the appendix. \ref{appendix:location} shows the exact location in the image where the secret invisible text is embedded.

\section{Conclusion}
In order to deploy deep neural networks in safety-critical areas and high-stakes applications, it's very important that the robustness evaluation is thorough, efficient, and covers the whole spectrum of semantically similar examples. Lately, there has been a huge surge in the amount of research in terms of creating adversarial attacks and defenses for deep neural networks, which has provided some great insights on evaluating the robustness. Though important, we argue that the current research in this field has not focused on simple unconventional methods on evaluating the robustness. This needs the attention because simple methods can be used any anyone to attack deep neural networks. Therefore, there's a high chance that attackers might adopt such simple and cheap methods to attack in the real-world and thus posing huge security concerns. Specifically, we propose simple transparent adversarial examples and illustrate their novel insights to complement current adversarial example generation pipeline on several image-based APIs. We hope that our unconventional route to highlight such simple attacks will motivate the research in this field to also consider the serious threats posed by such simple attacks to build more inclusive and broad robust defenses for machine learning systems in real-world settings. We also propose some potential future directions in \ref{appendix:future}.

\bibliography{iclr2021_conference}
\bibliographystyle{iclr2021_conference}



\newpage
\appendix
\section{Appendix}
\subsection{How are simple transparent adversarial examples different from the recently proposed simple transformations}
\label{appendix:simple}
Simple transformation such as translation and rotation \citep{rotation} and the recent non-programmatic Typographic attack \citep{goh2021multimodal} have been also found to fool deep neural networks. They may or may not require machine learning knowledge to design adversarial perturbations. But our attack method focuses on simple methods such as introducing transparent regions in the image and embedding invisible text. 

\subsection{simple transparent adversarial examples for object detection (more details)}
\subsubsection{Attack Algorithms to craft examples for object detection}
\label{appendix: attack1} 
\textbf{First approach}: for the first approach where we don't select the region to introduce transparent regions ourselves (i.e. let the tool decide it), we propose the following algorithm: 

\begin{itemize}
    \item \textbf{Step 1}: Using the transparency tool, we select some percentage of transparency to start with (10\%) and apply transparent patch on the image.  
    \item \textbf{Step 2}: We then query the object detection API with the transparently modified image. 
    \item \textbf{Step 3.1}: If the modified image evades object detection\footnote{We consider both cases where the image evades object detection or results in misclassification}, we then check for lesser transparency percentages (in this case < 10\%) till we get the image with the minimum transparency such that it evades the object detection but remains unambiguous and class-preserving to humans. 
    \item \textbf{Step 3.2}: If the transparently modified image doesn't evade the object detection, we keep increasing the transparency (in this case > 10\%) and querying the API till we get a modified sample such that it evades object detection but remains unambiguous and class-preserving to humans. We kept increasing the transparency by 5-10\% for the successive images in our experiments. 
\end{itemize}

\textbf{Second approach}
: for the second approach where we select specific region in the image to introduce transparent region, we propose the following algorithm:  
\begin{itemize}
    \item \textbf{Step 1}: We select any random region in the image to apply the transparent patch and we get the RGB color values of that region using any publicly available tool. 
    \item \textbf{Step 2}: We then select some percentage of transparency to start with (5\%) and provide those RGB color values to the transparency tool so that it would apply patch on that region(s) 
    \item \textbf{Step 3: }We then query the object detection API with the transparently modified image. 
    \item \textbf{Step 4.1}: If the modified image evades object detection, we then check for lesser transparency percentages (in this case < 5\%) till we get the image with the minimum transparency such that it evades the object detection but remains unambiguous and class-preserving to humans. 
     \item \textbf{Step 4.2}: If the transparently modified image doesn't evade the object detection, we keep increasing the transparency (in this case > 5\%) and querying the API till we get a modified sample such that it evades object detection but remains unambiguous and class-preserving to humans. We kept increasing the transparency by 1-2\% for the successive images in our experiments. 
\end{itemize}

\newpage
\subsubsection{crafting the example}
\label{appendix:craftingobj}
\begin{figure} [!hbt]  
  \begin{subfigure}{0.51\textwidth}
    \includegraphics[width=\linewidth]{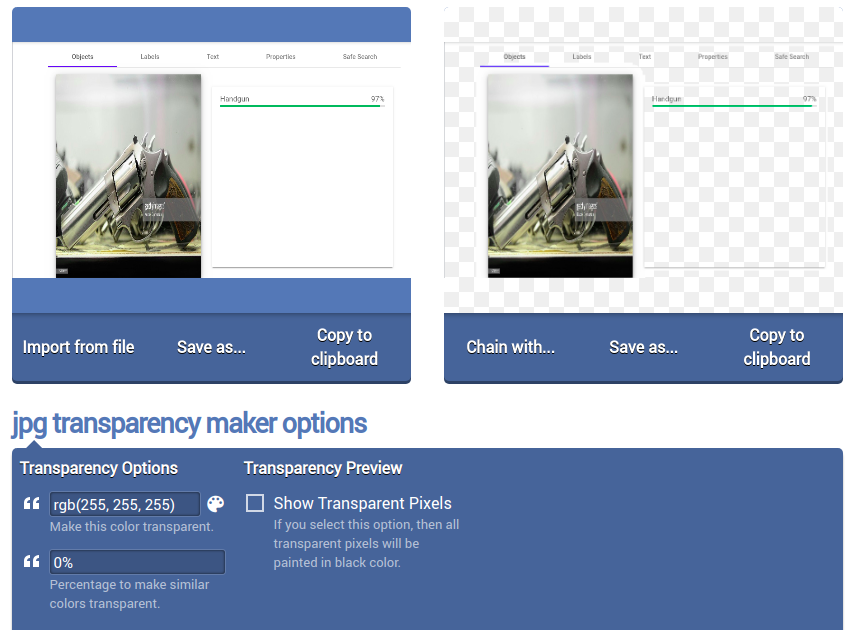}
    \caption{\textbf{Step 1:} Upload the image in the transparency tool}\vspace{0.5em} 
  \end{subfigure}\hspace{0.5em}%
  \begin{subfigure}{0.51\textwidth}
    \includegraphics[width=\linewidth]{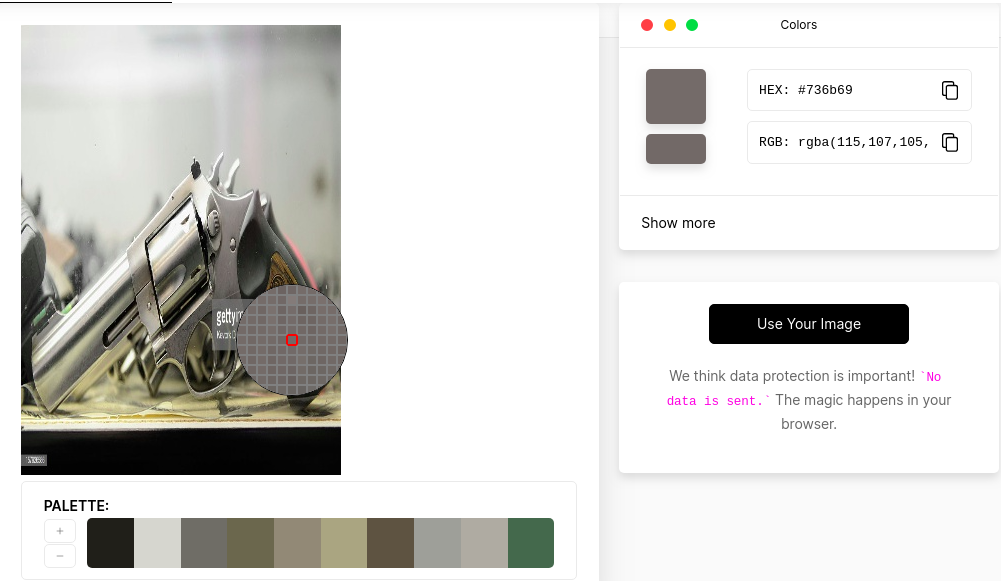}
    \caption{\textbf{Step 2:} Select the region in the image where you want to introduce transparent region. This can be done using any publicly available online tool.}\vspace{0.5em} 
  \end{subfigure}\hspace{0.5em} 
  \begin{subfigure}{0.51\textwidth}
    \includegraphics[width=\linewidth]{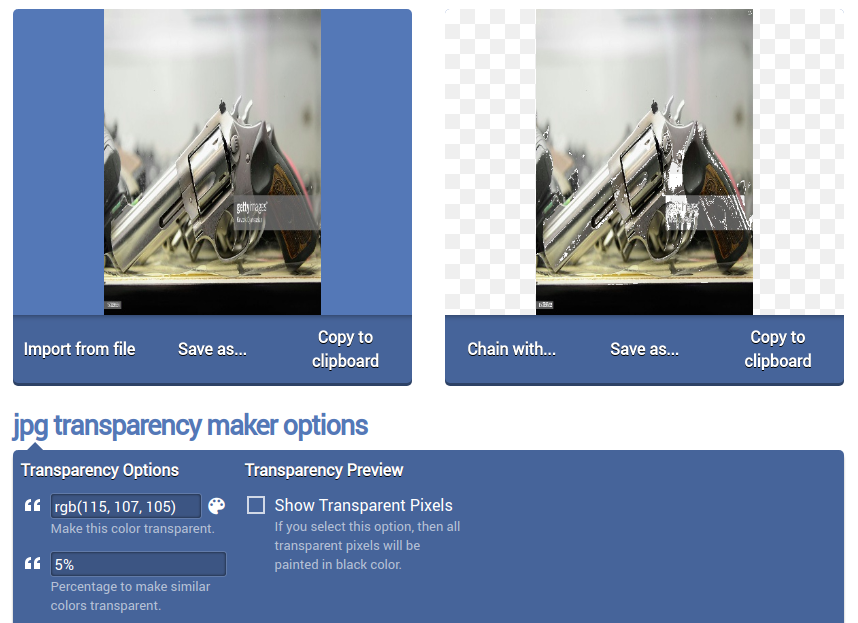}
    \caption{\textbf{Step 3}: Specify the region/color to be patched and transparency percentage in the tool. We start with 5$\%$ transparency.}\vspace{0.5em} 
  \end{subfigure}\hspace{0.5em}  
  \begin{subfigure}{0.51\textwidth}
    \includegraphics[width=\linewidth]{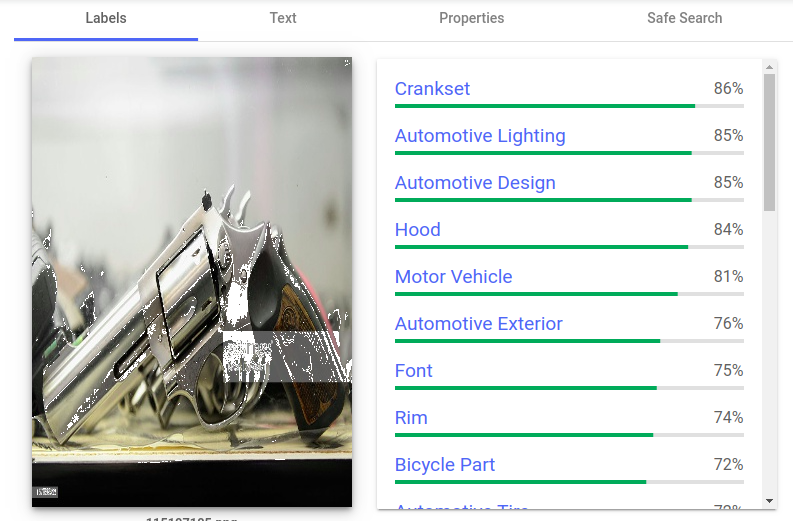}
    \caption{\textbf{Step 4:} Query the API}\vspace{0.5em} 
  \end{subfigure}\hspace{0.5em}
  \begin{subfigure}{0.51\textwidth}
    \includegraphics[width=\linewidth]{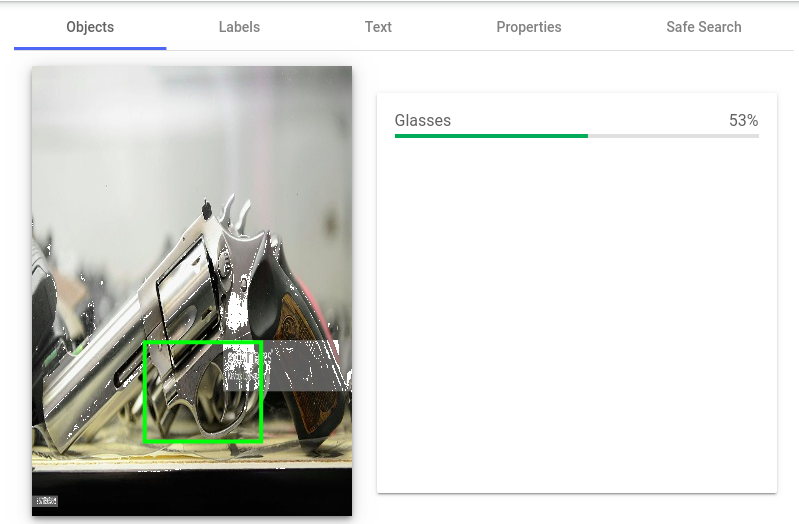}
    \caption{\textbf{Step 5:} Keep trying for lesser transparency. We try for 4$\%$ and it successfully fools the API. }\vspace{0.5em} 
  \end{subfigure}\hspace{0.5em}
  \begin{subfigure}{0.51\textwidth}
    \includegraphics[width=\linewidth]{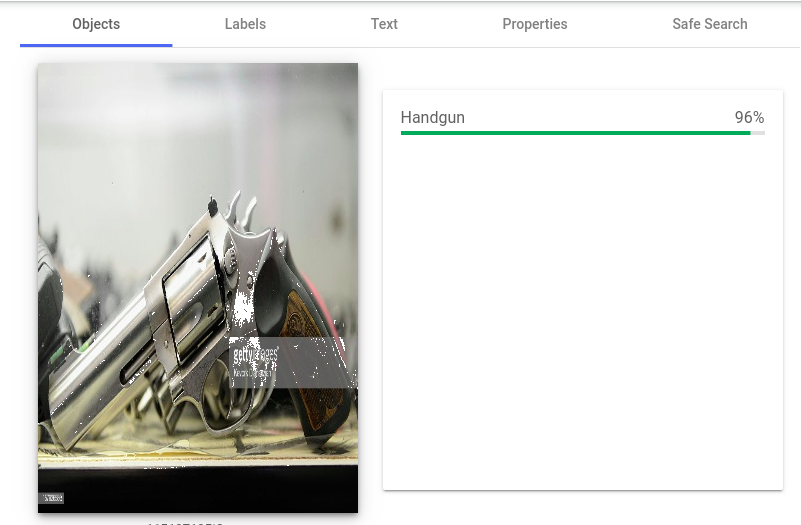}
    \caption{We further try for 3$\%$ transparency but it doesn't fool the API, so we stick with 4$\%$.}\vspace{0.5em} 
  \end{subfigure} 
  \caption{(a) denotes step 1 where the image to introduce transparent regions is uploaded in a transparency tool. One can use any publicly available similar transparency tool for this. (b) is step 2 where we use publicly available tool to select the region (RGB values) to introduce transparent regions. (c) denotes step 3 where we specify transparency intensity and location to introduce transparent region in the image. (d) We query the Google Cloud Vision API. As we can see, the modified image with transparent region fools the Google Cloud Vision API. In (e) and (f), we further check for lesser transparency intensity till we get the modified image such that it remains unambiguous to humans but fools the API.} \label{fig:craftobj}
\end{figure}

\subsubsection{Performance of object detection APIs against Simple Transparent Adversarial Examples}
\label{appendix: performance2}
\begin{table}[h!]
 \caption{Performance of object detection APIs against Simple Transparent Adversarial Examples. ASR denotes the attack success rate. Samples Tested show the number of modified images tested. Avg $\epsilon_1$ refers to the average modification constraint (i.e. transparency percentage) when we don't select any region ourselves to be patched and Avg $\epsilon_2$ refers to the average modification constraint when we select region(s) ourselves to be patched. Queries for $\epsilon_1$ and Queries for $\epsilon_2$ denote the average number of queries requires to fool the APIs when the modification constraints are $\epsilon_1$ and $\epsilon_2$. NA indicates that the modified images did not evade the object detection of that specific API.}  
\centering
\begin{adjustbox}{width=\textwidth}
\begin{tabular}{||c c c c c c c||} 
 \hline 
 API & ASR ($\%$) & Samples Tested & Avg $\epsilon_1$ & Queries for $\epsilon_1$ & Avg $\epsilon_2$ & Queries for $\epsilon_2$ \\ [0.5ex] 
 \hline\hline
 Azure Cloud Vision & 52$\%$ & 50 & 39.37$\%$ & 4 & 6.87$\%$ & 3 \\   
 Google Cloud Vision & 36$\%$ & 50 & 47.08$\%$ & 5 & 6.71$\%$ & 4 \\ 
 Sightengine & 9$\%$ & 50 & 43.25$\%$ & 4 & 6$\%$ & 2 \\ 
 Picpurify & 2$\%$ & 50 & 30$\%$ & 3 & NA & NA \\ [0.5ex] 
 \hline
\end{tabular}
\end{adjustbox}
\label{table:1}
\end{table}

Table \ref{table:1} summarizes the performance of all four APIs against simple transparent adversarial examples for object detection. Even though our attack has a lower attack success rate than the current norm-constrained  black-box attack methods, our attack has more realistic scenario and very likely to occur in real-world settings since it's very straightforward and query efficient to carry out. Picpurify and Sightengine APIs look more robust because they are specifically designed to detect weapons in images, whereas Azure Cloud Vision and Google Cloud Vision APIs are general object detection APIs that detect different types of objects in an image. We also observe that in the cases where attack is not successful, it does significantly degrade the accuracy. 

\subsubsection{Limitations}
One of the limitations of this attack is that if the object (for example a gun) to be patched in the image is very small than the overall size of the image, it might get difficult to apply a transparent patch on it. Even if the patch is applied, the object might look totally ambiguous to humans.   

\subsubsection{Possible risk of this attack}
Weapon detection APIs are deployed in various applications to filter out images containing weapons that might look violent and shocking or to simply detect weapons in images for high-stakes applications. Our results suggest potential risks that the attacker can very easily apply transparent patch to fool the weapon detection APIs, resulting in safety and security concerns.

\subsection{Simple Transparent Adversarial Examples for OCR (more details)}
\subsubsection{attack algorithm for ocr}
\label{appendix: ocrattackalgo}
\begin{itemize}
    \item \textbf{Step 1}: Based on our experiments, we start with a font size of 15 px and a difference of (at least) 10 each between the R, G, B values of the font color of the text and R, G, B values of the region where we want to embed the text. We call this difference as RGB difference\footnote{For example, if the R, G, B values of the font color of text are (100, 120, 200) and R, G, B values of the region are (110, 130, 210), then RGB difference will be 30}. We start with a minimum RGB difference of 30. Further, we test the embedded image on Google Cloud Vision API's OCR service.
    \item \textbf{Step 2.1}: If OCR is able to recognize the embedded text, we further keep reducing the \textit{RGB} difference or the font size or both up to the extent till which text can be recognized by OCR but evades the vision of time-limited humans.    
    \item \textbf{Step 2.2}: If OCR is not able to recognize the text, we steadily increase the \textit{RGB} difference keeping the font size as 15 px till the extent it gets recognized by OCR but evades the vision of time-limited humans.
   \end{itemize}
 
\newpage    
\subsubsection{crafting the example}
\label{appendix:craftingocr} 
\begin{figure} [!hbt]  
\centering
  \begin{subfigure}{0.45\textwidth}
    \includegraphics[width=\linewidth]{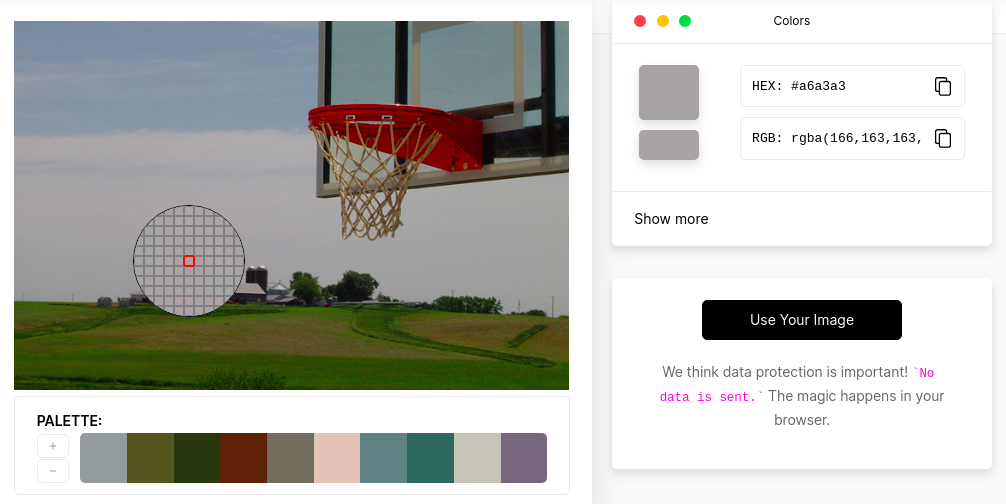}
    \caption{\textbf{Step 1} }\vspace{0.5em} 
  \end{subfigure}\hspace{0.7em} 
  \begin{subfigure}{0.45\textwidth}
    \includegraphics[width=\linewidth]{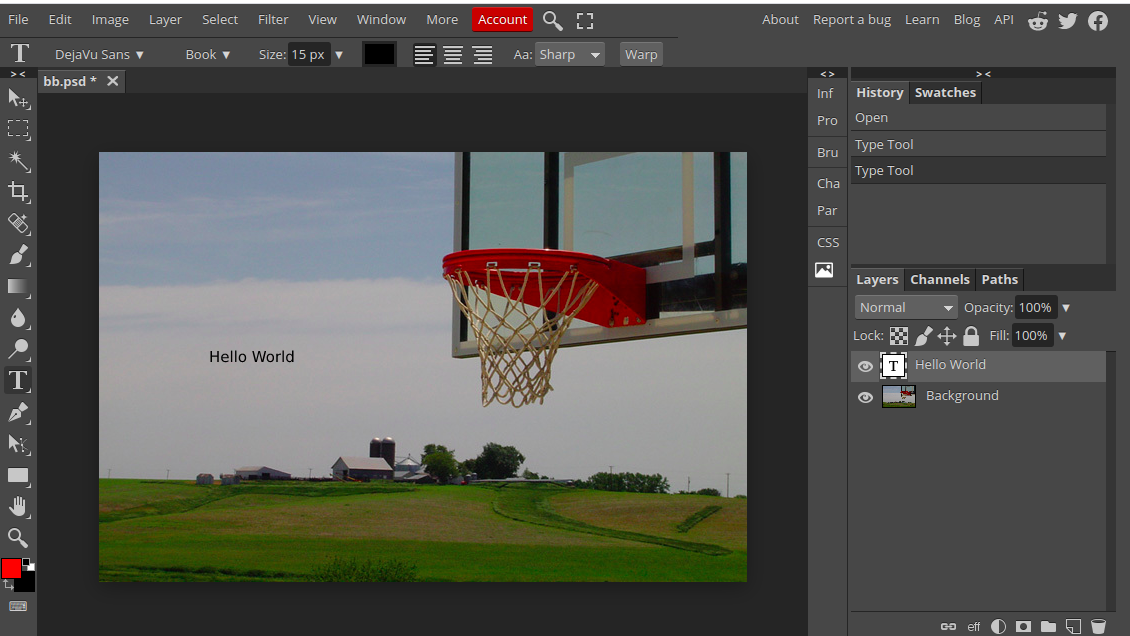}
    \caption{\textbf{Step 2} }\vspace{0.5em} 
  \end{subfigure} 
  \begin{subfigure}{0.45\textwidth}
    \includegraphics[width=\linewidth]{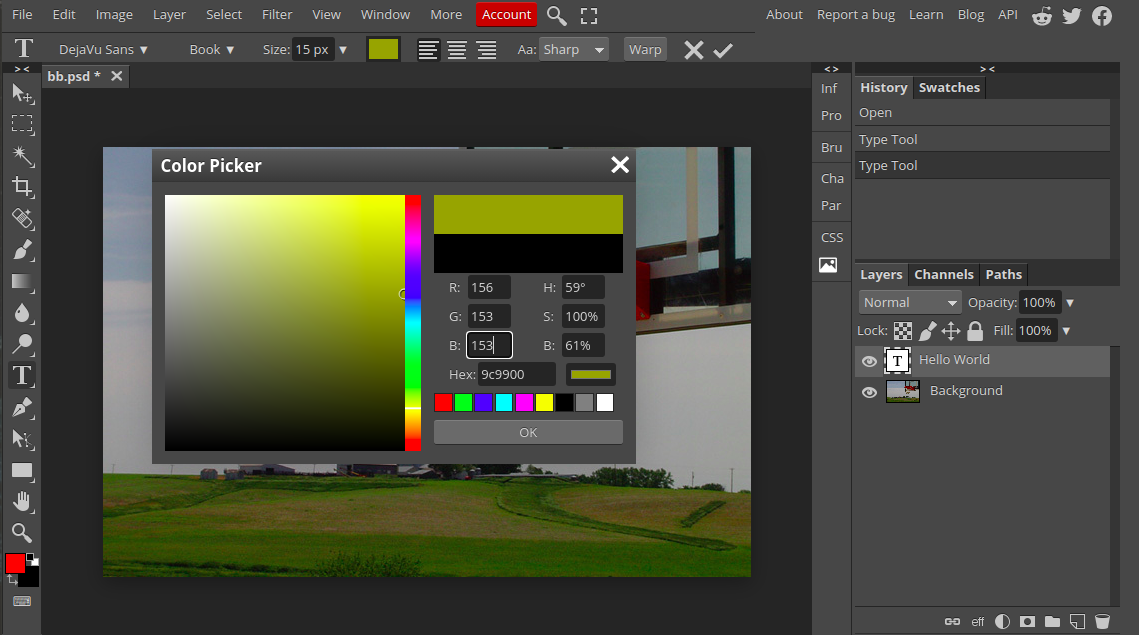}
    \caption{\textbf{Step 3} }\vspace{0.5em} 
  \end{subfigure}\hspace{0.3em}  
  \begin{subfigure}{0.45\textwidth}
    \includegraphics[width=\linewidth]{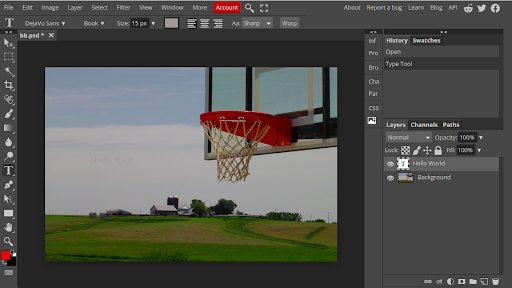}
    \caption{\textbf{Step 4}  }\vspace{0.5em} 
  \end{subfigure} 
  \begin{subfigure}{0.45\textwidth}
    \includegraphics[width=\linewidth]{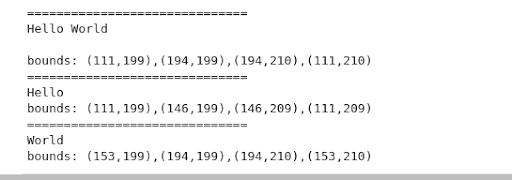}
    \caption{\textbf{Step 5}  }\vspace{0.5em} 
  \end{subfigure}\hspace{0.5em}
  \begin{subfigure}{0.45\textwidth}
    \includegraphics[width=\linewidth]{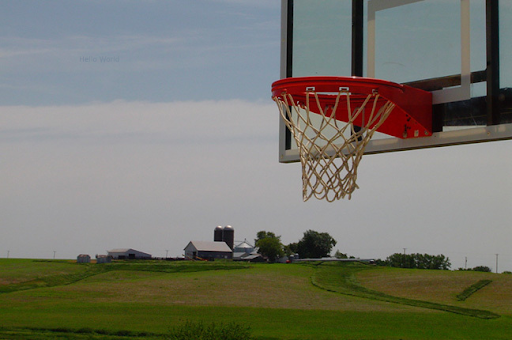}
    \caption{\textbf{Step 6} }\vspace{0.5em} 
  \end{subfigure}\hspace{0.5em}
   \begin{subfigure}{0.45\textwidth}
    \includegraphics[width=\linewidth]{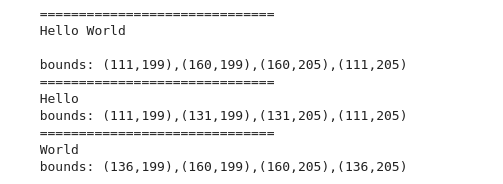}
    \caption{\textbf{Step 7} }\vspace{0.5em} 
  \end{subfigure}\hspace{0.5em}
   \begin{subfigure}{0.45\textwidth}
    \includegraphics[width=\linewidth]{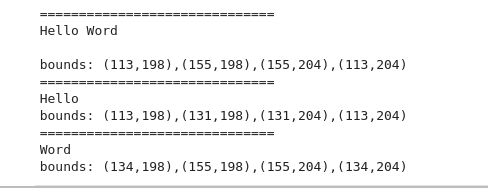}
    \caption{\textbf{Step 8}} 
  \end{subfigure}
  \caption{(a) shows step 1 where we select the region (RGB values) in the image to embed the text. Any publicly available tool can be used to do this. (b) In step 2, we embed the text "Hello World" with font size 15px using publicly available online editor. In (c) we set the RGB values of the text such that it gets invisible to human eyes. (d) shows the image with invisible text. Next in (e) we query the Google Cloud OCR and we can see that the Google Cloud OCR is able to detect the text that looks invisible to human eyes. (f) We keep checking for font sizes lesser than 15px to find the minimum font size. Here in this image we have the text with size 9px. (g) The Google Cloud OCR is still able to detect the text with font size 9px. (h) We further try for 8px but the Google Cloud OCR isn't able to detect it correctly. Hence, our ideal manipulated image will be the image with embedded text of size 9px.} \label{fig:craftocr}
\end{figure}

\newpage   
\subsubsection{Location of the secret text embedding} 
\label{appendix:location}
\begin{figure} [!hbt]  
  \begin{subfigure}{0.51\textwidth}
    \includegraphics[width=\linewidth]{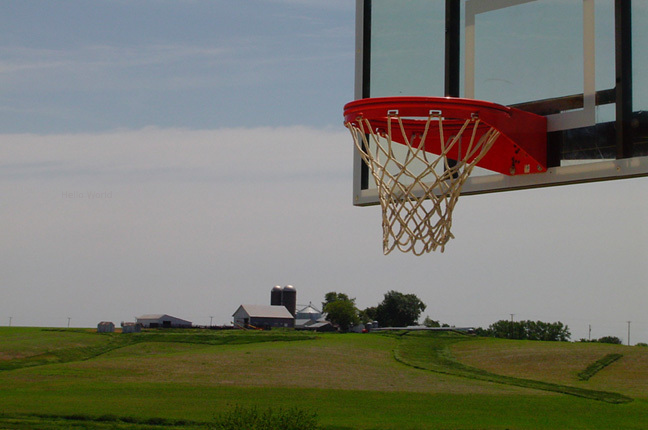}
    \caption{Modified image with invisible text embedding} 
  \end{subfigure}\hspace{0.5em}%
  \begin{subfigure}{0.51\textwidth}
    \includegraphics[width=\linewidth]{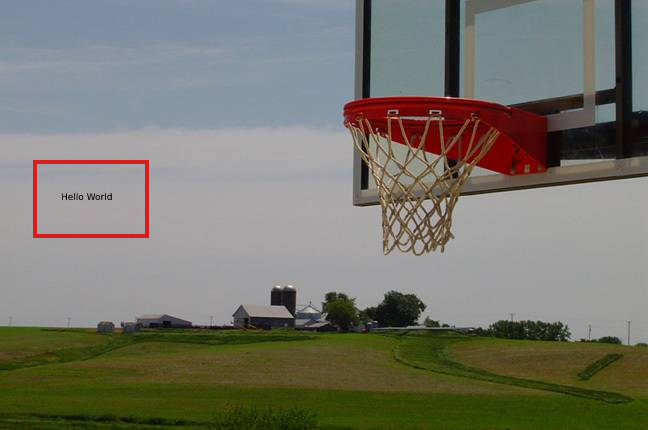}
    \caption{Modified image with black text embedding} 
  \end{subfigure} 
  \caption{(a) is the modified image that has secret invisible text of font size 9px and RGB values (160, 155, 157) in it and it evades the vision of time-limited humans but is detected by Google Cloud Vision API's OCR. To show exactly where it is placed, we change the font color to black and border it with a red rectangle that results in (b). The actual dimensions of the image are 648 px (width) and 430 px (height). } \label{fig:ocrloc}
\end{figure}

\subsubsection{Performance of Google Cloud Vision OCR against Simple Transparent Adversarial Examples}

\label{appendix: performance1}
\begin{table}[h!]
\caption{Performance of Google Cloud Vision OCR against Simple Transparent Adversarial Examples. RGB Values for Region denote the RGB values of the region where the secret text is to be embedded. Font color and Font size refer to the RGB values and size of the text to be embedded. Min RGB difference indicates the minimum possible difference between the RGB values of the region and RGB values of the text.}
\centering
\begin{tabular}{||c c c c||} 
 \hline 
 RGB Values for Region & Font color & Font size (px) & Min RGB difference \\ [0.5ex]
 \hline\hline
 
 (214, 44, 11) & (204, 34, 1) & 9 & 30 \\   
 (74, 0, 3) & (54, 0, 3) & 15 & 20 \\  
 (255, 255, 255) & (245, 245, 245) & 11 & 30 \\  
 (236, 236, 236) & (226, 226, 226) & 11 & 30 \\   
 (34, 73, 120) & (30, 63, 110) & 15 & 24 \\  
 (212, 235, 249) & (202, 225, 239) & 11 & 30 \\  
 (200, 202, 189) & (170, 182, 160) & 15 & 79 \\  
 (69, 37, 23) & (59, 27, 18) & 11 & 25 \\ 
 (126, 144, 162) & (120, 137, 155) & 9 & 20 \\ 
 (171, 170, 186) & (163, 162, 178) & 10 & 24 \\ [0.5ex] 
 \hline
\end{tabular} 
\label{table:ocrobs}
\end{table}

\subsubsection{Importance of the OCR attack}
\label{appendix: importance}
The current work on evaluating the robustness of deep neural networks based OCR systems revolves mostly around traditional adversarial examples \citep{song2018fooling, chen2020attacking}. We go beyond the traditional adversarial examples and introduce \textit{Simple Transparent Adversarial Examples} to evaluate the robustness of OCR systems. As vision APIs offering OCR services get deployed in high-stakes applications and safety-critical areas, such type of attacks can cause a tremendous loss. Moreover, as these type of adversarial images can be easily crafted by anyone with just online tools that are publicly available, they pose a higher security risk than the current typical adversarial examples. 
 
\subsubsection{Possible Applications and Risks of OCR attack}
\label{appendix: risks}
We show some ways in which the attacker can use Simple Transparent Adversarial Examples for OCR to attack high-stakes applications. 

\begin{enumerate}
    \item \textbf{Breaking Blind Review}: anonymized submissions can be broken by adding author names to the figures such that they evade the vision of time-limited humans but are detected by OCR applications. 
    \item \textbf{Breaking Check Scanner Systems}: A lot of check scanner APIs are available in the market that use OCR to process the checks. The attacker can embed a secret text, such as manipulating the amount or the name, which can have a major security threat. 
    \item \textbf{Steganography}: Steganography is the method of hiding secret data within a file to avoid detection.  \textit{Secret embedding approach} can be used as a simple technique to perform steganography, where image with secret invisible text can sent safely to its destination without the actual content being getting revealed.  
\end{enumerate}

\subsection{Images that fool both time-limited humans and Google Cloud Vision OCR} 
\begin{figure}[hbt!] 
  \begin{subfigure}{0.23\textwidth}
    \includegraphics[width=\linewidth]{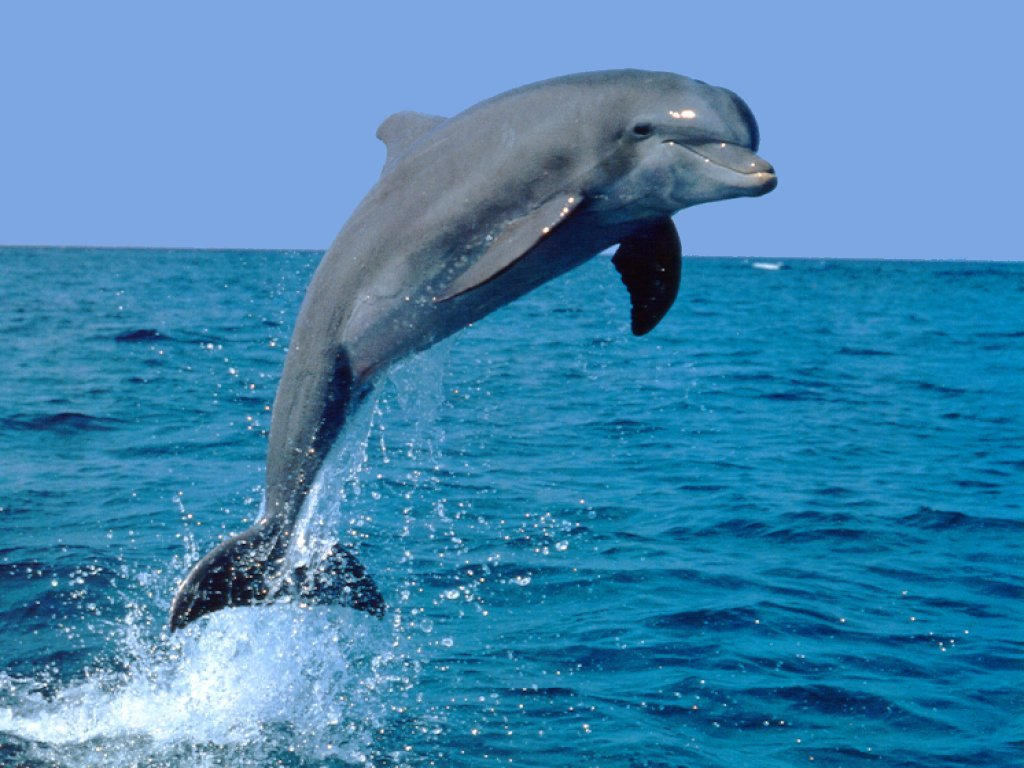}
    \caption{Modified image} \label{fig:2b}
  \end{subfigure}\hspace{0.5em}%
  \hspace*{\fill}   
  \begin{subfigure}{0.23\textwidth}
    \includegraphics[width=\linewidth]{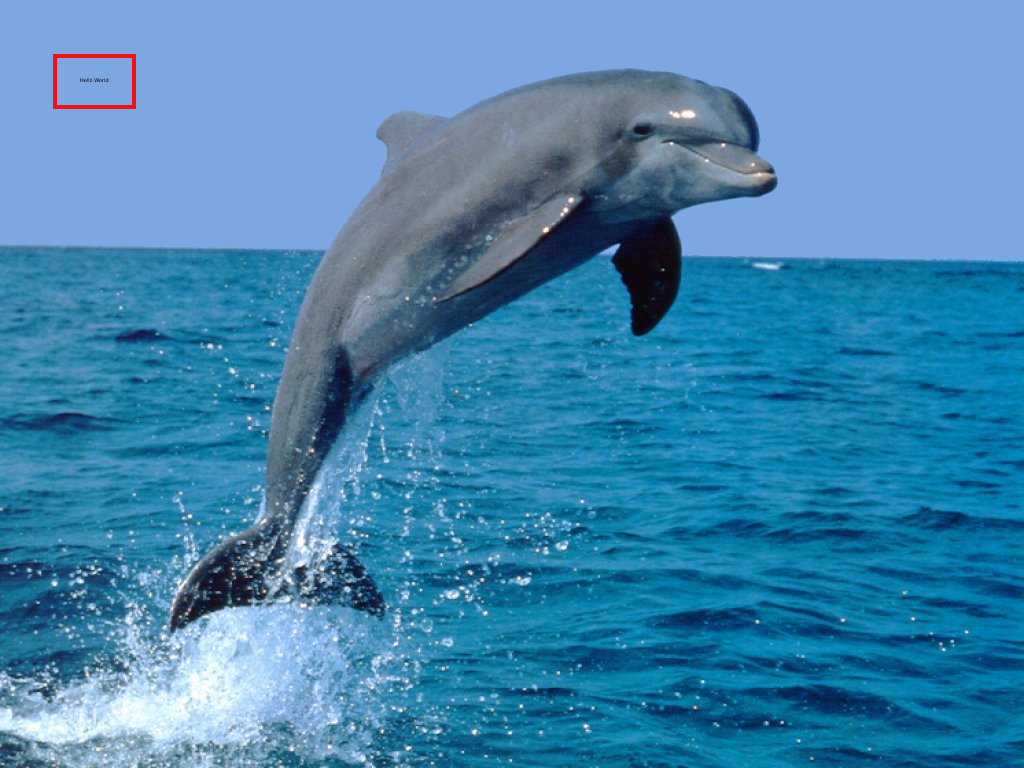}
    \caption{Modified image with black text} \label{fig:2c}
  \end{subfigure}\hspace{0.5em}%
  \hspace*{\fill} 
   \begin{subfigure}{0.23\textwidth}
    \includegraphics[width=\linewidth]{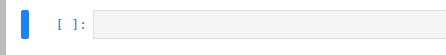}
    \caption{Google Cloud OCR output} \label{fig:2d}
  \end{subfigure}
  \caption{(a) is the image that has secret embedded text "Hello World" inside it of font size 5px and RGB difference set to 0 such that it evades the vision of both time-limited humans and Google Cloud OCR. (b) shows the same embedded text but in black color to give the idea about where the text is embedded in (a). (c) demonstrates the OCR's output that shows that the embedded text also evades OCR's vision. } \label{fig:foolboth}
\end{figure}

Figure \ref{fig:foolboth} demonstrates the examples where the image with the secretly embedded text fools the vision of both time-limited humans and Google Cloud's OCR. 
\label{appendix:foolboth}

\subsection{venn diagram}
\begin{figure}[htp]
    \centering
    \includegraphics[width=9cm]{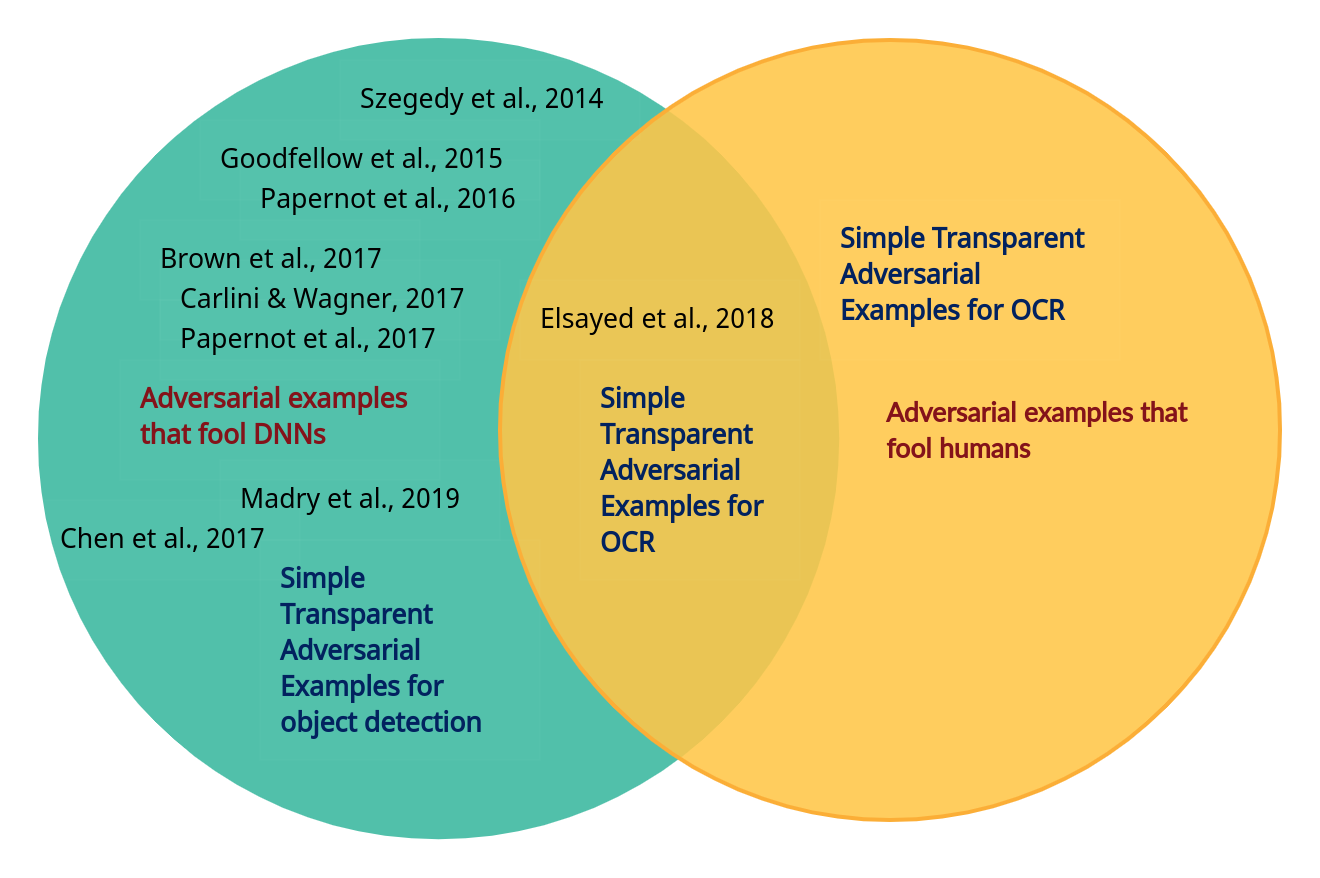}
    \caption{The left circle in the venn diagram represents the adversarial attack methods that fool only deep neural networks. The right circle represents the attacks that fool only time-limited humans. And the intersection represents the attack methods that fool both deep neural networks and time-limited humans. In case of object detection, simple transparent adversarial examples fool only deep neural networks. But in case of OCR, they fool only time-limited humans (and not Google Cloud OCR), as well as both the vision of time-limited humans and deep neural networks powered Google Cloud OCR at the same time when the RGB difference and font size of the text are set very small.}
    \label{fig:venn}
\end{figure}

Venn Diagram in Figure \ref{fig:venn} demonstrates the evasive nature of our attack when compared with the conventional adversarial attack methods. 

\subsection{Future directions}
\label{appendix:future} 
There are also some interesting future directions. First, can we effectively apply the transparent patch on other objects apart from the guns such that they still remain unambiguous to humans but evade object detection? Second, it would be also interesting to see how robust are adversarially trained models to these simple attacks. Our primary focus has been on evaluating the robustness of publicly deployed APIs since the attacker can very easily query them through the web interface in the real-world. Third, it's also worthwhile to explore what type of current defenses can be potentially used against this attack. 

\newpage
\subsection{More results}
\label{appendix:moreresults}
\begin{figure}[hbt!] 
  \begin{subfigure}{0.10\textwidth}
    \includegraphics[width=\linewidth]{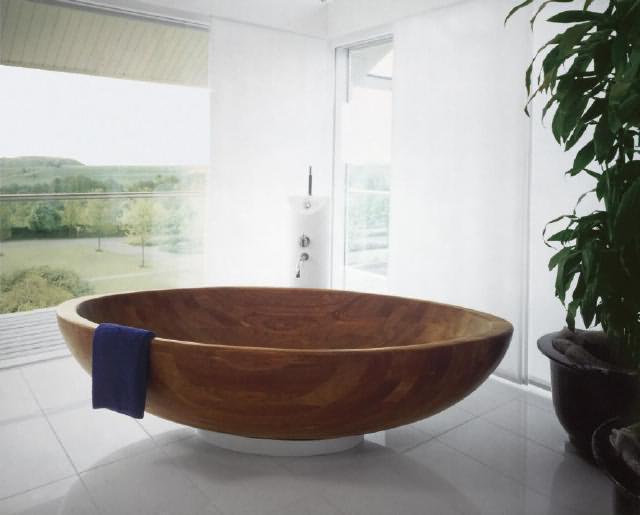}
    \caption{ } 
  \end{subfigure}\hspace{0.3em}%
  \begin{subfigure}{0.1\textwidth}
    \includegraphics[width=\linewidth]{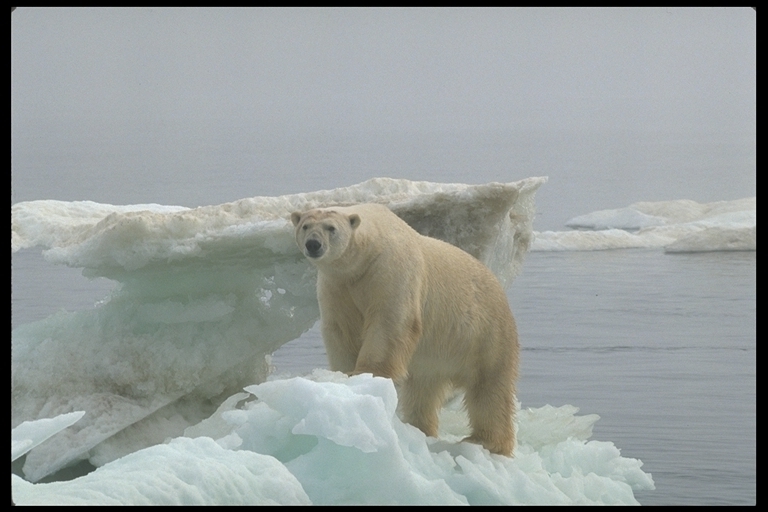}
    \caption{ } 
  \end{subfigure}\hspace{0.3em}%
  \begin{subfigure}{0.1\textwidth}
    \includegraphics[width=\linewidth]{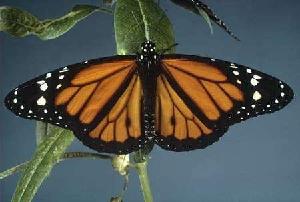}
    \caption{ } 
  \end{subfigure}\hspace{0.3em}%
   \begin{subfigure}{0.1\textwidth}
    \includegraphics[width=\linewidth]{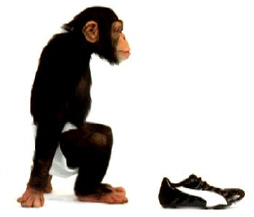}
    \caption{ } 
  \end{subfigure}\hspace{0.3em}%
   \begin{subfigure}{0.1\textwidth}
    \includegraphics[width=\linewidth]{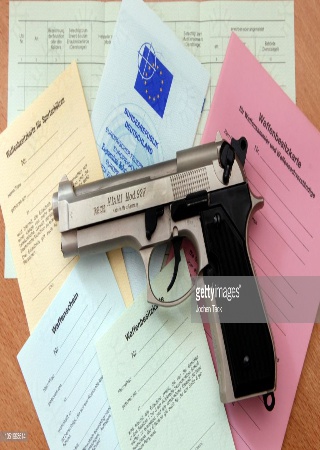}
    \caption{ } 
  \end{subfigure}\hspace{0.3em}%
   \begin{subfigure}{0.1\textwidth}
    \includegraphics[width=\linewidth]{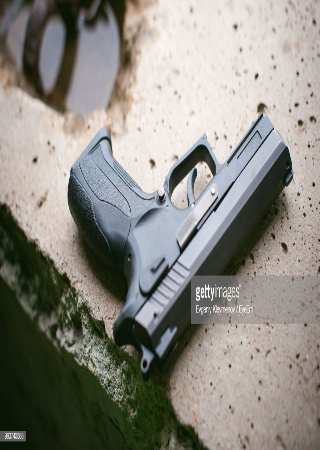}
    \caption{ } 
  \end{subfigure}\hspace{0.3em}%
   \begin{subfigure}{0.1\textwidth}
   \includegraphics[width=\linewidth]{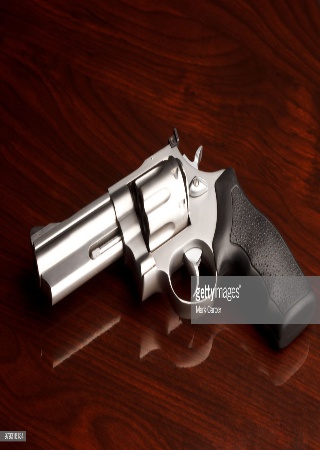}
    \caption{ } 
  \end{subfigure}\hspace{0.3em}%
   \begin{subfigure}{0.1\textwidth}
    \includegraphics[width=\linewidth]{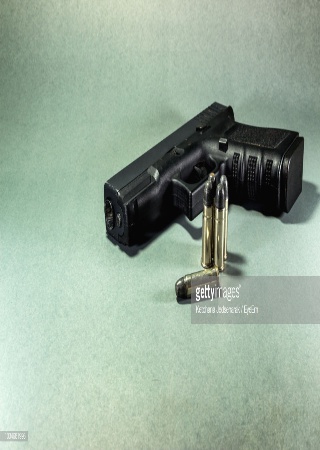}
    \caption{ } 
  \end{subfigure}\hspace{0.3em}%
   \begin{subfigure}{0.1\textwidth}
    \includegraphics[width=\linewidth]{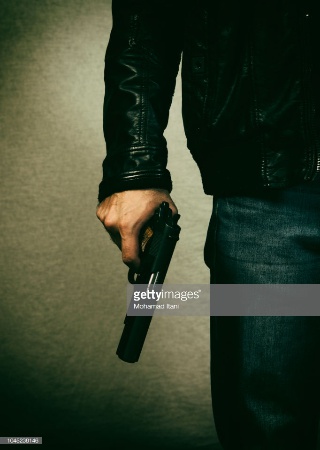}
    \caption{ } 
  \end{subfigure}
  
  \begin{subfigure}{0.10\textwidth}
    \includegraphics[width=\linewidth]{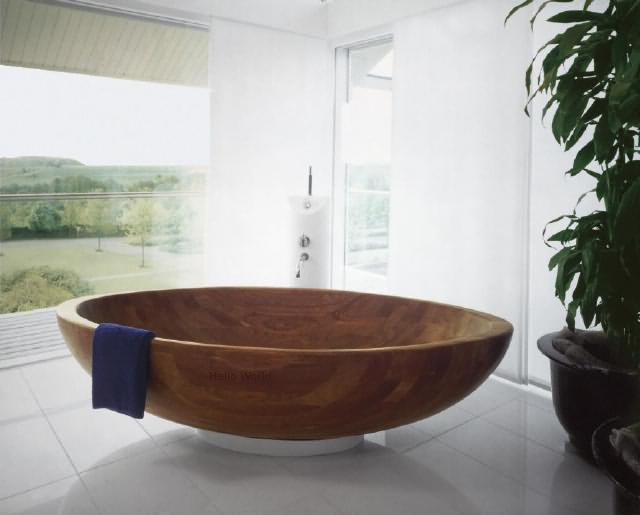}
    \caption{ } 
  \end{subfigure}\hspace{0.3em}%
  \begin{subfigure}{0.1\textwidth}
    \includegraphics[width=\linewidth]{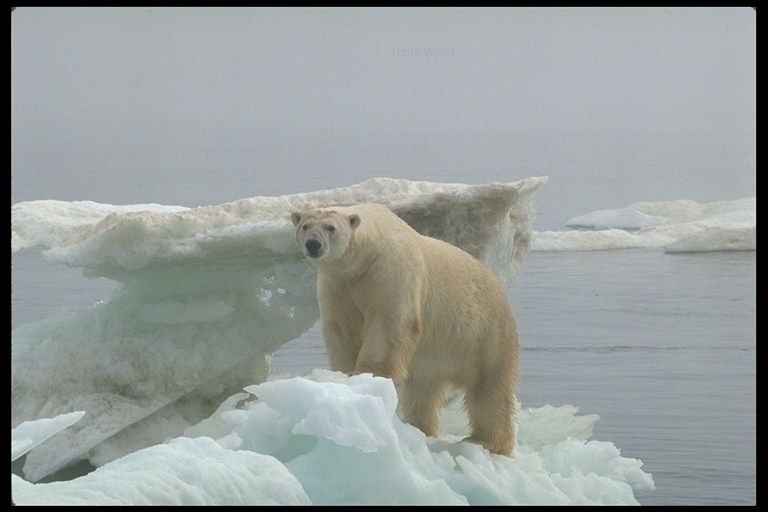}
    \caption{ } 
  \end{subfigure}\hspace{0.3em}%
  \begin{subfigure}{0.1\textwidth}
    \includegraphics[width=\linewidth]{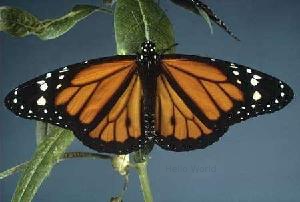}
    \caption{ } 
  \end{subfigure}\hspace{0.3em}%
   \begin{subfigure}{0.1\textwidth}
    \includegraphics[width=\linewidth]{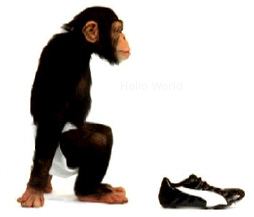}
    \caption{ } 
  \end{subfigure}\hspace{0.3em}%
   \begin{subfigure}{0.1\textwidth}
    \includegraphics[width=\linewidth]{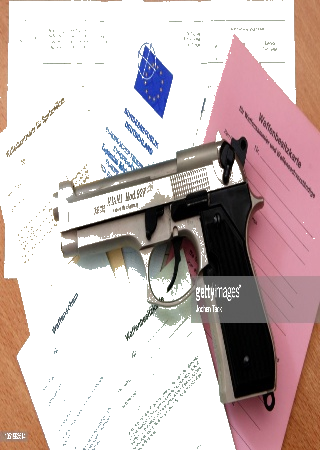}
    \caption{ } 
  \end{subfigure}\hspace{0.3em}%
   \begin{subfigure}{0.1\textwidth}
    \includegraphics[width=\linewidth]{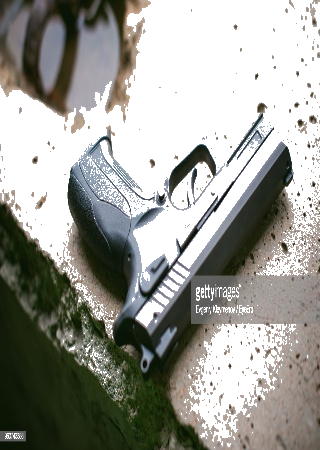}
    \caption{ } 
  \end{subfigure}\hspace{0.3em}%
   \begin{subfigure}{0.1\textwidth}
    \includegraphics[width=\linewidth]{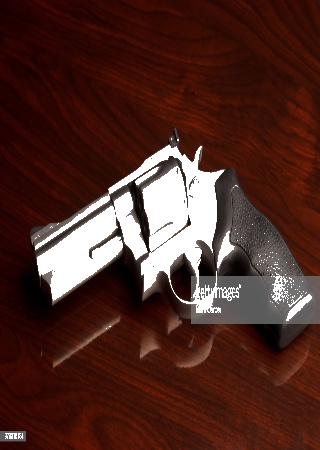}
    \caption{ } 
  \end{subfigure}\hspace{0.3em}%
   \begin{subfigure}{0.1\textwidth}
    \includegraphics[width=\linewidth]{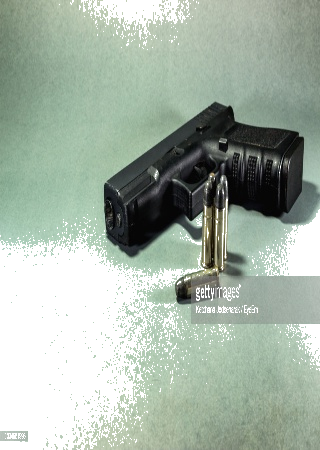}
    \caption{ } 
  \end{subfigure}\hspace{0.3em}%
   \begin{subfigure}{0.1\textwidth}
    \includegraphics[width=\linewidth]{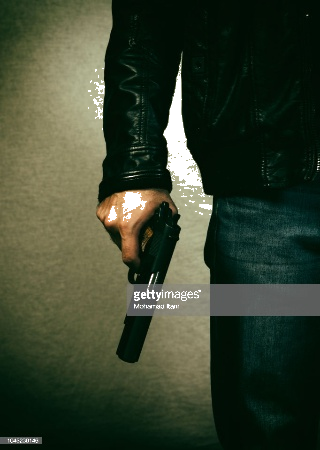}
    \caption{ } 
  \end{subfigure}
\caption{(a) to (i) are original unmodified images. (j) to (m) are modified by adding secret embedding "Hello World" with varying font sizes and font colors. They evade vision of time-limited humans but are detected by Google Cloud Vision's OCR. (n) to (r) are modified by adding white transparent patches using an online tool. They fool object detection APIs such as sightengine.com, PicPurify, Google Cloud Vision, and Microsoft Azure's computer vision API either by evading detection or getting misclassified. (p), (q), and (o) are misclassified as packaged goods, camera, and grooming trimmer by Google Cloud Vision API, (n), (o), and (r) evade sightengine.com, (o) and (r) evade PicPurify. } \label{fig:7}
\end{figure}   

\newpage
\begin{figure} [!hbt]  
   \begin{subfigure}{0.51\textwidth}
    \includegraphics[width=\linewidth]{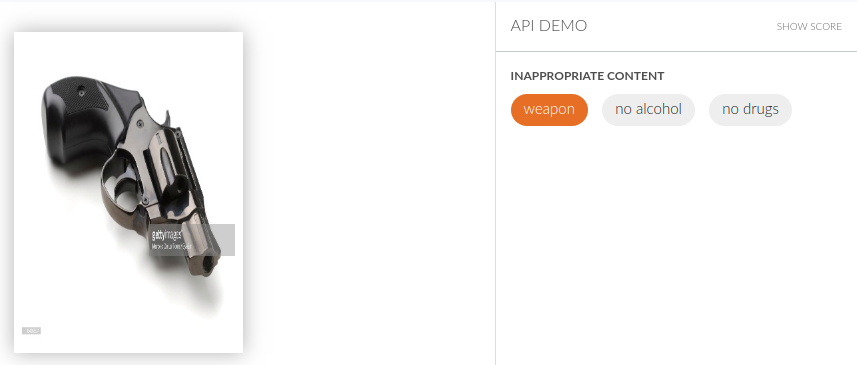}
    \caption{Original image} 
  \end{subfigure}\hspace{0.5em}%
  \begin{subfigure}{0.51\textwidth}
    \includegraphics[width=\linewidth]{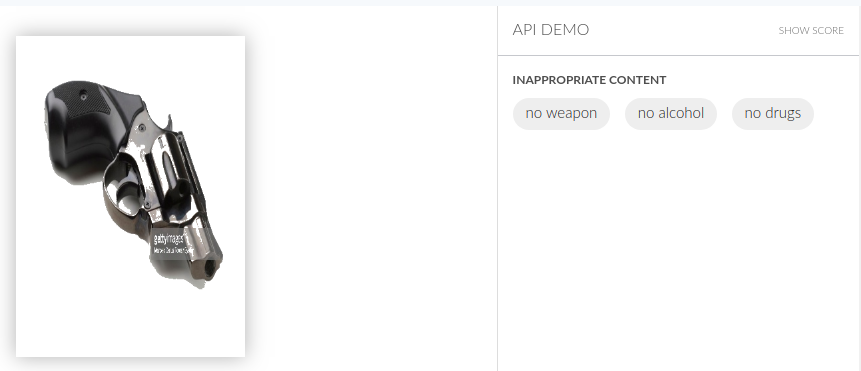}
    \caption{Modified image} 
  \end{subfigure} 
  \caption{(a) is the original image having gun that is detected successfully by Sightengine API. (b) is the modified image having 28$\%$ transparency intensity that evades the Sightengine API's weapon detection.} \label{fig:8}
\end{figure}
\end{document}